\documentclass[journal]{IEEEtran}
\usepackage{threeparttable}
\usepackage[numbers,sort&compress]{natbib}

\ifCLASSINFOpdf
  \usepackage[pdftex]{graphicx}
  \graphicspath{{./imgs/}}
  \DeclareGraphicsExtensions{.pdf,.jpeg,.png}
\else
\fi
\usepackage{amsmath, mathrsfs, eucal, amssymb}
\usepackage{algorithm, algorithmic}

\usepackage{array, booktabs, makecell}
\usepackage{url, hyperref}

\begin{document}
%
\title{A Semantics-Assisted Video Captioning Model Trained with Scheduled Sampling}
%
%
%

\author{Haoran~Chen,
        Ke~Lin,
        Alexander~Maye,
        Jianming~Li, 
        and~Xiaolin~Hu
\thanks{Haoran Chen was from and Jianmin Li, Xiaolin Hu are from Department of Computer Science and Technology, Tsinghua University, Beijing 100084, China. 
Email: \href{mailto:chr17@tsinghua.org.cn}{chr17@tsinghua.org.cn}, \href{mailto:lijianmin@tsinghua.edu.cn}{lijianmin@tsinghua.edu.cn}, \href{mailto:xlhu@tsinghua.edu.cn}{xlhu@tsinghua.edu.cn}}
\thanks{Ke Lin is from Samsung Research China, Beijing (SRC-B), China. Email: \href{mailto:ke17.lin@samsung.com}{ke17.lin@samsung.com}}
\thanks{Alexander Maye is from Department of Neurophysiology and Pathophysiology, University Medical Center, Hamburg-Eppendorf, Hamburg, Germany.
Email: \href{mailto:a.maye@uke.de}{a.maye@uke.de}}}

%
%

\markboth{Semantics-Assisted Video Captioning}%
{Semantics-Assisted Video Captioning}
%



\maketitle

\begin{abstract}
Given the features of a video, recurrent neural networks can be used to automatically generate a caption for the video. Existing methods for video captioning have at least three limitations. First, semantic information has been widely applied to boost the performance of video captioning models, but existing networks often fail to provide meaningful semantic features. Second, the Teacher Forcing algorithm is often utilized to optimize video captioning models, but during training and inference, different strategies are applied to guide word generation, leading to poor performance. Third, current video captioning models are prone to generate relatively short captions that express video contents inappropriately. Toward resolving these three problems, we suggest three corresponding improvements. First of all, we propose a metric to compare the quality of semantic features, and utilize appropriate features as input for a semantic detection network (SDN) with adequate complexity in order to generate meaningful semantic features for videos. Then, we apply a scheduled sampling strategy that gradually transfers the training phase from a teacher-guided manner toward a more self-teaching manner. Finally, the ordinary logarithm probability loss function is leveraged by sentence length so that the inclination of generating short sentences is alleviated. Our model achieves better results than previous models on the YouTube2Text dataset and is competitive with the previous best model on the MSR-VTT dataset.
\end{abstract}


\IEEEpeerreviewmaketitle

\section{Introduction}
Video captioning aims to automatically generate a concise and accurate description for a video. 
It requires techniques both from computer vision (CV) and natural language processing (NLP). 
Deep learning (DL) methods for sequence-to-sequence learning are able to learn the map from discrete color arrays to dense vectors, 
which is utilized to generate natural language sequences without the interference of humans. 
These methods produced impressive results on this task compared with the results yielded by manually crafted features.

It has gained increasing attention in video captioning that the semantic meaning of a video is critical and beneficial for an RNN to generate annotations \cite{pan2016jointly, gan2017semantic}. 
Keeping semantic consistency between video content and video description helps to refine a generated sentence in semantic richness \cite{gao2017video}. 
But few researches have explored methods to obtain video semantic features, metrics to measure their quality and the relation between video captioning performance and meaningfulness of semantic features.

Several training strategies have been used to optimize video captioning models, such as the Teacher Forcing algorithm and CIDEnt-RL \cite{pasunuru2017reinforced}. 
The Teacher Forcing algorithm is a simple and intuitive way to train RNNs. 
But it suffers from the discrepancy between training, which utilizes ground truth to guide word generation at each step, and inference, which samples from the model itself at each step. 
Reinforcement learning (RL) techniques have also been adopted to improve the training process of video captioning. 
CIDEnt-RL is one of the best RL algorithms, but it is extremely time-consuming to calculate metrics for every batch. 
In addition, the improvement on different metrics is unbalanced. 
In other words, the improvements on other metrics are not as large as that on the specific metrics optimized directly.

The commonly used loss function for video captioning is comprised of the logarithm of probabilities of target correct words \cite{venugopalan2015sequence, donahue2015long-term}. 
A long sentence tends to bring high loss to the model, 
as each additional word reduces the joint probability by roughly at least one order of magnitude. 
In contrast, a short sentence with few words has a relatively low loss. 
Thus, a video captioning model is prone to generate short sentences after being optimized by a log likelihood loss function. 
Excessively short annotations may neither be able to describe a video accurately nor express the content of a video in a rich language. 

We propose to improve solutions to the video captioning task in three aspects. 
Firstly, we use mean average precision (mAP) as the metric to evaluate the quality of semantic information. 
By virtue of the evaluation metric, 
we build our semantic detection network (SDN) with a proper scale and the best inputs that brings the best performance, 
and, consequently, SDN is able to produce meaningful and accurate semantic features for a video. 
Secondly, we take advantage of a scheduled sampling method to train our video captioning model, 
which searches extreme points in the RNN state space more extensively as well as bridges the gap between training process and inference \cite{bengio2015scheduled}. 
Thirdly, we optimize our model by a sentence-length-modulated loss function, which encourages the model to generate longer captions with more detail.

Our implementation, available on GitHub\footnote{\url{https://github.com/WingsBrokenAngel/Semantics-AssistedVideoCaptioning/tree/master}}, is based on the TensorFlow deep learning framework.
\section{Related Works}
\subsection{Image Captioning}
The encoder-decoder paradigm has been widely applied by researchers in image captioning since it was introduced to machine translation \cite{cho2014learning}. 
It has become a mainstream method in both image captioning and machine translation \cite{vinyals2015show, mao2014explain}. 
Inspired by successful attempts to employ attention in machine translation \cite{bahdanau2015neural} and object detection \cite{ba2015multiple}, 
models that are able to attend to key elements in an image are investigated for the purpose of generating high-quality image annotations. 
Semantic features \cite{you2016image} and object features \cite{anderson2018bottom-up} are incorporated into attention mechanisms as heuristic information to guide selective and dynamic attendance of salient segments in images. 
RL techniques, which optimize specific metrics of a model directly, are also adopted to enhance the performance of image captioning models \cite{rennie2017self-critical}. 
Graph Convolutional Networks (GCNs) have been introduced to cooperate with RNN to integrate both semantic 
and spatial information into image encoders in order to generate efficient representations of an image \cite{yao2018exploring}. 
Stimulated by the success of the Transformer model in machine translation, researchers extend it to a multimodal model for image captioning\cite{yu2019multimodal}, 
which utilizes multi-view visual features to further improve the performance. 
Multi-level relationships between image regions are learnt and both low- and high-level features are exploited at the decoding stage 
in the Meshed Transformer with memory for image captioning\cite{2019arXiv191208226C}.

\subsection{Video Captioning}
Though both image captioning and video captioning are multi-modal tasks, video captioning is probably harder than the former one, as videos show not only spatial features but also temporal correlations.

Following the successful adoption of the encoder-decoder paradigm in image captioning, 
multimodal features of videos are fed into a sequence-to-sequence model to generate video descriptions 
with the assistance of pretrained models in image classification \cite{venugopalan2015sequence, donahue2015long-term}. 
In order to alleviate the semantic inconsistency between the video content and the generated caption, 
visual features and semantic features of a video are mapped to a common embedding space 
so that semantic consistency may be achieved 
by minimizing the Euclidean distance between these two embedded features \cite{pan2016jointly}. 
A model named POS generates video captions with Part-of-Speech (POS) information 
and multiple representations of video clips\cite{Wang_2019_ICCV}. 
MARN exploits a memory structure to explore the relation between a word 
and its various visual contexts across the training data\cite{Pei_2019_CVPR}. 
JSRL-VCT manages to generate video descriptions by corporating visual representations and syntax representations\cite{Hou_2019_ICCV}. 
GRU-EVE captures rich temporal dynamics in video features by Short Fourier Transform, 
and extracts semantic information from an object detector\cite{Aafaq_2019_CVPR}. 
\cite{Zheng_2020_CVPR} proposes a Syntax-Aware Action Targeting (SAAT) component to learn an action and its subjects that exist in a video for better semantic consistency in captioning. 

RNN, especially LSTM, can be extended by integrating high-level tags or attributes of video with visual features of the video through embedding and element-wise addition/multiplication operations \cite{gan2017semantic}. 
\cite{yu2016video} exploits a sentence generator that is built upon an RNN module to model language, 
a multimodal layer to integrate different modal information, and an attention module to dynamically select salient features from the input. 
The output of a sentence generator is fed into a paragraph generator for describing a relatively long video with several sentences. 

Following the attention mechanism introduced by \cite{xu2015show}, \cite{gao2017video} captures the salient structure of video with the help of visual features of the video and context information provided by LSTM.  
Although bottom-up \cite{anderson2018bottom-up} and top-down attention \cite{ramanishka2017top-down} have been proposed for image captioning, selectively focusing on salient regions in an image is, 
to some extent, similar to picking key frames in a video \cite{chen2018less}. 
\cite{wang2018watch} explores crossmodal attention at different granularity levels and capture global temporal structures as well as local temporal structures implied in multimodal features to assist the generation of video captions. 

Due to the lack of labeled video data and the abundance of unlabeled video data, 
\cite{pasunuru2017multi-task} and \cite{sun2019videobert} propose to improve video captioning with self-supervised learning tasks 
or unsupervised learning tasks, such as unsupervised video prediction, entailment generation and text-to-video generation. 
\cite{pasunuru2017multi-task} demonstrates that multi-task training contributes to sharing knowledge across different domains, 
and each task, including video captioning, benefits from the training of other irrelevant tasks. 
\cite{sun2019videobert} takes advantage of the abundance of unlabeled videos on YouTube 
and train the BERT model introduced in \cite{devlin2018bert} on comparably large-scale videos, 
which is then used as a feature extractor for video captioning. 
A large amount of pre-training data is critical to BERT models both 
in video captioning and machine translation \cite{devlin2018bert,sun2019videobert}.
By aggregating different experts on different known activities, \cite{wang2019learning} takes advantage of external textual corpora and transfer knowledge to unseen data for zero-shot video captioning. 
A spatio-temporal graph model is built to find object interactions and knowledge distillation mechanism is proposed to increase stability of performance\cite{Pan_2020_CVPR}. 

\subsection{RNN Training Strategy}
The traditional method to train an RNN is the Teacher Forcing algorithm \cite{williams1989a}, 
which feeds human annotations to the RNN as input at each step to guide the token generation during training and samples a token from the model itself as input during inference. 
The different sources of input tokens during training and inference lead to the inability of the model to generate high-quality tokens in inference, as errors may accumulate along the sequence generation.

\cite{bengio2015scheduled} proposes to switch gradually from guiding generation by true tokens to feeding sampled tokens during training, 
which helps RNN models adapt to the inference scheme in advance. 
It has been applied to image captioning and speech recognition. 
Inspired by \cite{huszar2015how}, who mathematically proves that both the Teacher Forcing algorithm and Curriculum Learning have a tendency to learn a biased model, 
\cite{lamb2016professor} solves the problem by adopting an adversarial domain method to align the dynamics of the RNN during training and inference.
\cite{Zhang_2020_CVPR} proposes an object relational graph (ORG) to encode interaction features and design a teacher-recommended learning (TRL) method to utilize linguistic knowledge. 

Inspired by the successful application of RL methods in image captioning \cite{rennie2017self-critical}, 
\cite{pasunuru2017reinforced} proposes a modified reward that compensates for the logical contradiction in phrase-matching metrics as the direct optimization target in video captioning. 
The gradient of the non-differentiable RL loss function is computed and back-propagated by the REINFORCEMENT algorithm \cite{williams1992simple}. 
But calculation of the reward for each training batch adds a non-negligible computation cost to the training process and slows down the optimization progress. 
In addition, the improvements of RL methods on various metrics are not comparable with the improvement on the specific metric used as RL reward.  

\section{The Proposed Approaches}
We consider the video captioning task as a supervised task. 
The training set is annotated as $N$ pairs of $\{\mathbf{X}_i, \hat{\mathbf{Y}}_i\}$, 
where $\mathbf{X}_i$ denotes a video and $\hat{\mathbf{Y}}_i$ represents the corresponding target caption. 
Suppose there are $M$ frames from a video and a caption consisting of $L_i$ words, then we have:
\begin{equation}
\begin{aligned}
\mathbf{X}_i&=\{\boldsymbol{x}_{i,0}, \boldsymbol{x}_{i,1}, \dots, \boldsymbol{x}_{i, M-1}\},\\
\hat{\mathbf{Y}}_i&=\{\hat{\boldsymbol{y}}_{i,0}, \hat{\boldsymbol{y}}_{i,1}, \dots, \hat{\boldsymbol{y}}_{i,L_i-1}\}, 
\label{eq:1}
\end{aligned}
\end{equation}
where each $\boldsymbol{x}$ denotes a single frame and each $\boldsymbol{y}$ denotes a word belonging to a fixed known dictionary. 

A pretrained model is used to produce word embeddings, and we obtain a low-dimension embedding of the caption $\hat{\mathbf{Y}}_i \in \mathbb{R}^{L_i\times D_w} $,
\begin{equation}
\hat{\mathbf{Y}}_i = (\boldsymbol{w}_{i,0}, \boldsymbol{w}_{i,1}, \dots, \boldsymbol{w}_{i, L_i-1})^T, 
\quad \boldsymbol{w}_{i,j} \in \mathbb{R}^{D_w},
\end{equation}
where $D_w$ is the dimension of the word embedding space.  
 
\subsection{Encoder-Decoder Paradigm}
\subsubsection{Encoder}
Our encoder is composed of a 3D ConvNet, a 2D ConvNet and a semantic detection network (SDN). 
The 3D ConvNet is utilized to produce the spatio-temporal feature $\boldsymbol{e}_i \in \mathbb{R}^{D_e}$ for the $i$th video. 
The 2D ConvNet is supposed to find the static visual feature $\boldsymbol{r}_i \in \mathbb{R}^{D_r}$ for the $i$th video. 
The visual spatio-temporal representation of the $i$th video can then be obtained by concatenating both features together: 
\begin{equation}
\boldsymbol{v}_i = \begin{pmatrix}\boldsymbol{r}_i \\ \boldsymbol{e}_i\end{pmatrix} \in \mathbb{R}^{D_v}, 
\label{eq:2}
\end{equation}
where $D_v=D_e+D_r$.

For semantic detection, we manually select the $K$ most common and meaningful words, 
which consists of the most frequent nouns, verbs or adjectives, 
from both the training set and the validation set as candidate tags for all videos \cite{gan2017semantic}. 
The semantic detection task is treated as a multi-label classification task 
with $\boldsymbol{v}_i$ as the representation of the $i$th video 
and $\hat{\boldsymbol{s}}_i=\{\hat{s}_{i,0}, \hat{s}_{i,1}, \dots, \hat{s}_{i,K-1}\} \in \{0,1\}^{K}$ as the ground truth. 
If the $j$th tag exists in the annotations of the $i$th video, then $\hat{s}_{i,j}=1$; otherwise, $\hat{s}_{i,j}=0$. 
Suppose $\boldsymbol{s}_i$ is the semantic feature of the $i$th video. 
Then, we have $\boldsymbol{s}_i=\sigma(f(\boldsymbol{v}_i))\in (0,1)^{K}$, 
where $f(\cdot)$ is a nonlinear mapping and $\sigma(\cdot)$ a sigmoid activation function. 
Mean average precision is applied to evalute the quality of semantic features. 
A multi-layer perceptron (MLP) of adequate scale is exploited to learn semantic representations from the samples. 
The set of input features is determined by the experimental results for each dataset. 
The SDN is trained by minimizing the loss function: 
\begin{equation}
\begin{aligned}
L(\boldsymbol{s}_i, \hat{\boldsymbol{s}}_i)=& \\
\frac{1}{N}\sum_{i=0}^{N-1}& \sum_{j=0}^{K-1} \hat{s}_{i,j}\log{s_{i,j}}+(1-\hat{s}_{i,j})\log{(1-s_{i,j})}. 
\end{aligned}
\label{eq:3}
\end{equation}
A probability distribution of tags $\boldsymbol{s}_i$ is produced by the SDN to 
represent the semantic content of the $i$th video in the training set, the validation set or the test set. 

\subsubsection{Decoder}
Standard RNNs \cite{elman1990finding} are capable of learning temporal patterns from input sequences. 
But they suffer from the gradient vanishing/explosion problem, which results in their inability to generalize to long sequences. 
LSTM \cite{hochreiter1997long} is a prevailing variant of RNN that alleviates the long-term dependency problem by using gates to update the cell state, 
but it ignores the semantic information of the input sequence. 
We use SCN(Semantic Compositional Network) \cite{gan2017semantic}, a variant of LSTM, as our decoder, 
because it not only avoids the long-term dependency problem but also takes advantage of semantic information of the input video. 
Suppose we have a video feature $\boldsymbol{v}$, a semantic feature $\boldsymbol{s}$, an input vector $\boldsymbol{x}_{t}$ at time step $t$ and a hidden state $\boldsymbol{h}_{t-1}$ at time step $t-1$ . 
The SCN integrates semantic information $\boldsymbol{s}$ into $\boldsymbol{v}$, $\boldsymbol{x}_{t}$ and $\boldsymbol{h}_{t-1}$, respectively, 
and obtains the semantics-related video feature $\hat{\boldsymbol{v}}$, the semantics-related input $\hat{\boldsymbol{x}}_{t}$ 
and the semantics-related hidden state $\hat{\boldsymbol{h}}_{t-1}$ as follows: 
\begin{equation}
\begin{aligned}
\hat{\boldsymbol{x}}_{z,t} &= \mathbf{W}_{z,c} \cdot ((\mathbf{W}_{z,a} \cdot \boldsymbol{x}_{t}) \odot  (\mathbf{W}_{z,b} \cdot \boldsymbol{s})), \\
\hat{\boldsymbol{v}}_{z}&=\mathbf{C}_{z,c} \cdot ((\mathbf{C}_{z,a} \cdot \boldsymbol{v}) \odot (\mathbf{C}_{z,b} \cdot \boldsymbol{s})), \\
\hat{\boldsymbol{h}}_{z, t-1} &= \mathbf{U}_{z,c} \cdot ((\mathbf{U}_{z,a} \cdot \boldsymbol{h}_{t-1}) \odot (\mathbf{U}_{z,b} \cdot \boldsymbol{s})), \\
&\quad z \in \{c, i, f, o\},
\label{eq:4}
\end{aligned}
\end{equation}
where $c$, $i$, $f$ and $o$ denote the cell state, the input gate, the forget gate and the output gate, respectively. 

Then input gate $\boldsymbol{i}_t$, forget gate $\boldsymbol{f}_t$ and output gate $\boldsymbol{o}_t$ at time step $t$ are calculated, respectively, in a way similar to the standard LSTM as follows: 
\begin{equation}
\begin{aligned}
\boldsymbol{i}_t &= \sigma(\hat{\boldsymbol{x}}_{i,t}+\hat{\boldsymbol{h}}_{i,t-1}+\hat{\boldsymbol{v}}_{i}+\boldsymbol{b}_{i}), \\
\boldsymbol{f}_t &= \sigma(\hat{\boldsymbol{x}}_{f,t}+\hat{\boldsymbol{h}}_{f,t-1}+\hat{\boldsymbol{v}}_{f}+\boldsymbol{b}_{f}), \\
\boldsymbol{o}_t &= \sigma(\hat{\boldsymbol{x}}_{o,t}+\hat{\boldsymbol{h}}_{o,t-1}+\hat{\boldsymbol{v}}_{o}+\boldsymbol{b}_{o}),
\label{eq:5}
\end{aligned}
\end{equation}
where $\sigma$ denotes the logic sigmoid function $\sigma(x) = \frac{1}{1+e^{-x}} \in (0, 1)$ and $\boldsymbol{b}$ is a bias term for each gate. 

The raw cell state at the current step $t$ can be computed as follows: 
\begin{equation}
\hat{\boldsymbol{c}}_t = 
\tanh{(\hat{\boldsymbol{x}}_{c,t}+\hat{\boldsymbol{h}}_{c,t-1}+\hat{\boldsymbol{v}}_c+\boldsymbol{b}_c)}, \label{eq:6}
\end{equation}
where $\tanh$ denotes the hyperbolic function $\tanh{(x)}=\frac{e^x-e^{-x}}{e^{x}+e^{-x}}\in (-1, 1)$ and $\boldsymbol{b}_c$ is the bias term for the cell state. 
The input gate $\boldsymbol{i}_{t}$ is supposed to control the throughput of the semantic-related input $\hat{\boldsymbol{x}}_t$, 
and the forget gate $\boldsymbol{f}_t$ is designed to determine the preservation of the previous cell state $\boldsymbol{c}_{t-1}$. Thus, we have the final cell state $\boldsymbol{c}_t$ at time step: 
\begin{equation}
\boldsymbol{c}_t = \boldsymbol{f}_t * \boldsymbol{c}_{t-1} + \boldsymbol{i}_t * \hat{\boldsymbol{c}}_t. \label{eq:7}
\end{equation}

The output gate controls the throughput ratio of the cell state $\boldsymbol{c}_{t}$ so that the cell output $\boldsymbol{h}_t$ can be determined by: 
\begin{equation}
\boldsymbol{h}_t = \boldsymbol{o}_t * \tanh{(\boldsymbol{c}_t)}. \label{eq:8}
\end{equation}

The semantics-related variables $\hat{\boldsymbol{x}}_{t}$, $\hat{\boldsymbol{v}}$, $\hat{\boldsymbol{h}}_{t-1}$ and $\hat{\boldsymbol{c}}_t$ are dependent on semantic feature $\boldsymbol{s}$ 
so that the SCN takes semantic information of the video into account implicitly. 
The forget gate $\boldsymbol{f}_t$ is a key component in updating $\boldsymbol{c}_{t-1}$ to $\boldsymbol{c}_t$, which, to some degree, avoids the long-term dependency problem. 

\subsection{Training Method}
In the context of the RNN trained with the Teacher Forcing algorithm, the logarithmic probability $P(Y_i|X_i; \Theta)$ of a given triplet of input/output/label $(X_i, Y_i, \hat{Y}_i)$ and given model parameters $\Theta$ can be calculated as: 
\begin{equation}\label{rnnprob}
P(Y_i|X_i; \Theta) = \sum_{t=0}^{L_i-1}\log P(y_{i,t}|\hat{y}_{i,0},\cdots, \hat{y}_{i,t-1}, X_i;\Theta),
\end{equation}
where $L_i$ is the length of output. 

In the case of SCN, the joint logarithmic probability can be computed as: 
\begin{equation}
\begin{aligned}
P(Y_i|X_i; \Theta)& \\
= \sum_{t=0}^{L_i-1}&\log P(y_{i,t}|\hat{y}_{i,0}, \cdots, \hat{y}_{i, t-1}, s_i,  X_i; \Theta),\\
=\sum_{t=0}^{L_i-1}&\log P(y_{i,t}|h_{i, t-1}, c_{i, t-1}, \hat{y}_{i,t-1}, s_i, X_i; \Theta),
\end{aligned}
\end{equation}
where $h_{i,t}$, $c_{i,t}$ and $s_i$ are the output state, the cell state and the semantic feature of the $i$th video, respectively.

To some extent, $h_{i, t}$ and $c_{i,t}$ can be viewed as the aggregation of all the previous information. 
We can compute them using the recurrence relation: 
\begin{equation}\label{eq:9}
\begin{aligned}
h_{i,t} =\left\{
\begin{array}{rl}
 f(X_i, h_{i,t-1}, c_{i,t-1}, s_i, X_i; \Theta)  & \text{if   } t=0,\\
 f(\hat{y}_{i,t-1}, h_{i,t-1}, c_{i,t-1}, s_i,X_i; \Theta) & \text{if   } t>0,
 \end{array}\right. \\
 c_{i,t} = \left \{
 \begin{array}{rl}
 g(X_i, h_{i,t-1}, c_{i,t-1}, s_i, X_i; \Theta) & \text{if   } t=0, \\
 g(\hat{y}_{i,t-1}, h_{i, t-1}, c_{i, t-1}, s_i, X_i; \Theta) & \text{if   } t>0,
 \end{array}\right. 
\end{aligned}
\end{equation}
where $h_{i, -1} = \mathbf{0}$ , $c_{i,-1}=\mathbf{0}$. In inference, we need to replace $\hat{y}_{i,t}$ with $y_{i,t}$, which may lead to the accumulation of prediction errors.

In order to bridge the gap between training and testing in the Teacher Forcing algorithm, we train our video captioning model with scheduled sampling. 
Scheduled sampling transfers the training process gradually from using ground truth words $\hat{Y}_i$ for guiding to using sampled words $Y_i$ for guiding at each recurrent step. 
The commonly used strategy to sample a word from the output distribution is $\arg\max$. 
But the search scope is limited to a relatively small part of the search space, since it always selects the word with the largest probability. 
For the sake of enlarging the search scope, we draw a word randomly from the output distribution as a part of the input for the next recurrent step. 
In this way, words with higher probabilities are more likely to be chosen. 
The randomness of the sampling procedure will enable the recurrent network to explore a relatively large range of the network state space. 
In addition, the network is less likely to get stuck in a local minimum. 
In the perspective of training machine learning models, the multinomial sampling strategy reduces overfitting of the network; in other words, it acts like a regularizer.

Our method to optimize the language model consists of two parts: 
the outer loop schedule the sampling probability at each recurrent step (Algorithm \ref{alg:schedule}), 
while the algorithm inside the RNN (Algorithm \ref{alg:sample}) specifies the procedure to sample from the output of a model with a given possibility as a part of the input for the next step of the RNN.

\begin{algorithm} [htb]
	\caption{Scheduling Algorithm: schedule the $\epsilon$ across epochs}
	\label{alg:schedule}
	\begin{algorithmic}[1]
		\REQUIRE $EPOCH$: max epoch number, $STEPS\_PER\_EPOCH$: steps per epoch, $\mathbf{feature}$: necessary features
		\STATE ${\epsilon}list \leftarrow \textit{generate\_epsilon}()$  \COMMENT{Generate $epsilon$ for each epoch by a predeterminate strategy.}
		\STATE $\mathbf{output} \leftarrow \mathbf{0}$
		\FOR{$i=0$ \TO $EPOCH$}
			\FOR{$j=0$ \TO $STEPS\_PER\_EPOCH$}
				\STATE $\mathbf{output}_{i,j} \leftarrow \textit{function}(\mathbf{feature}_{i,j}, {\epsilon}list[i])$  \COMMENT{Run RNN}
				\STATE optimize the network with an optimizer
				\STATE extend $\mathbf{output}$ with $\mathbf{output}_{i,j}$
			\ENDFOR
		\ENDFOR
		\RETURN $\mathbf{output}$
	\end{algorithmic}
\end{algorithm}

\begin{algorithm}[htb]
	\caption{Random Sampling Algorithm: specific procedures in RNN}
	\label{alg:sample}
	\begin{algorithmic}[1]
		\REQUIRE $\mathbf{v}_i$: video feature, $\mathbf{s}_i$: semantic feature, $\mathbf{x}_i$: input array, $\epsilon$: sampling probability, $STEP$: max time step
		\ENSURE $\mathbf{h}_i$: output state, $\mathbf{c}_i$: cell state
		\STATE $\mathbf{h}_{i,0} \leftarrow \mathbf{0}$
		\STATE $\mathbf{c}_{i,0} \leftarrow \mathbf{0}$
		\STATE $\mathbf{h}_{i} \leftarrow \mathbf{0}$
		\STATE $\mathbf{c}_{i} \leftarrow \mathbf{0}$
		\STATE $\mathbf{embed} \leftarrow \mathbf{x}_{i,0}$
		\FOR{$t=1$ \TO $STEP$}
		\STATE $\mathbf{h}_{i,t}, \mathbf{c}_{i,t} \leftarrow \textit{recurrent\_step}(\mathbf{h}_{i, t-1}, \mathbf{c}_{i,t-1}, \mathbf{v}_i, \mathbf{s}_i, \mathbf{embed})$
		\STATE extend $\mathbf{h}_i$ with $\mathbf{h}_{i,t}$
		\STATE extend $\mathbf{c}_i$ with $\mathbf{c}_{i,t}$
		\STATE $prob \leftarrow \textit{random}(0, 1)$
		\IF{$prob < \epsilon$}
		\STATE $\mathbf{prob\_dist}_{i,t} \leftarrow \textit{word\_dist\_map}(\mathbf{h}_{i,t})$ \COMMENT{Map output state to word probability.}
		\STATE $\mathbf{word\_index} \leftarrow \textit{multinomial}(\mathbf{prob\_dist}_{i,t})$ \COMMENT{Sample from the word distribution.}
		\STATE $\mathbf{embed} \leftarrow \textit{lookup\_embed}(\mathbf{word\_index})$ \COMMENT{Use an embedding vector to represent the word.}
		\ELSE
		\STATE $\mathbf{embed} \leftarrow \mathbf{x}_{i,t}$
		\ENDIF
		\STATE $t \leftarrow t + 1$
		\ENDFOR
		\RETURN $\mathbf{h}_i, \mathbf{c}_i$
	\end{algorithmic}
\end{algorithm}

\subsection{Sentence-Length-Related Loss Function}
What is a good description for a video? 
A good description should be both accurate and concise. 
In order to achieve this goal, we design a sentence-length-modulated loss function \eqref{eq:loss1} for our model. 
\begin{equation}
\begin{aligned}
&\mathbf{Loss}(\hat{y}_{i}, s_i, X_i; \Theta) =\\ 
&-\sum_{i=0}^{b_s-1}\frac{1}{L_i^{\beta}} \sum_{t=0}^{L_i-1}\log p(\hat{y}_{i,t}|h_{i,t-1}, c_{i,t-1}, s_i, X_i; \Theta),
\end{aligned}
\label{eq:loss1}
\end{equation}
where $b_s$ is the batch size and $\beta>=0$ is a hyper-parameter 
that is used to keep a balance between the conciseness and the accuracy of the generated captions. 
If $\beta=0$, \eqref{eq:loss2} is a loss function commonly used in video captioning tasks. 
\begin{equation}
\begin{aligned}
&\mathbf{Loss}(\hat{y}_i, s_i, X_i; \Theta) =\\
&-\sum_{i=0}^{b_s-1}\sum_{t=0}^{L_i-1}\log p(\hat{y}_{i,t}|h_{i,t-1}, c_{i,t-1}, s_i, X_i; \Theta).
 \end{aligned}
\label{eq:loss2}
\end{equation}
In this loss function, a long sentence has greater loss than a short sentence. 
Thus, after minimizing the loss, 
the RNN is inclined to generate relatively short annotations that may be incomplete in semantics or sentence structure. 
If $\beta=1$, all words in the generated captions are treated equally in the loss function as well as in the process of optimization, 
which may lead to redundancy or duplicate words in the process of generating captions.

Thus, we have the following optimization problem:
\begin{equation}
\begin{aligned}
&\Theta = \\ 
&\arg \min_{\Theta} -\sum_{i=0}^{N-1}\frac{1}{L_i^{\beta}} 
\sum_{t=0}^{L_i-1}\log p(\hat{y}_{i,t}|h_{i,t-1}, c_{i,t-1}, s_i, X_i; \Theta),
\end{aligned}
\end{equation} 
where $N$ is the size of the training data and $\Theta$ is the parameter of our model.

GNMT, Google's Neural Machine Translation system, employs a similar length-normalization technique in the beam search during test, 
but not during training\cite{2016arXiv160908144W}. 
In contrast, our model abandons beam search in the decoder, 
and the model parameters are optimized by the sentence-length-modulated loss function\eqref{eq:loss1}. 
Note that beam search makes the decoding process slower. 

\begin{figure*}[htb]
\begin{center}
\includegraphics[width=0.8\textwidth]{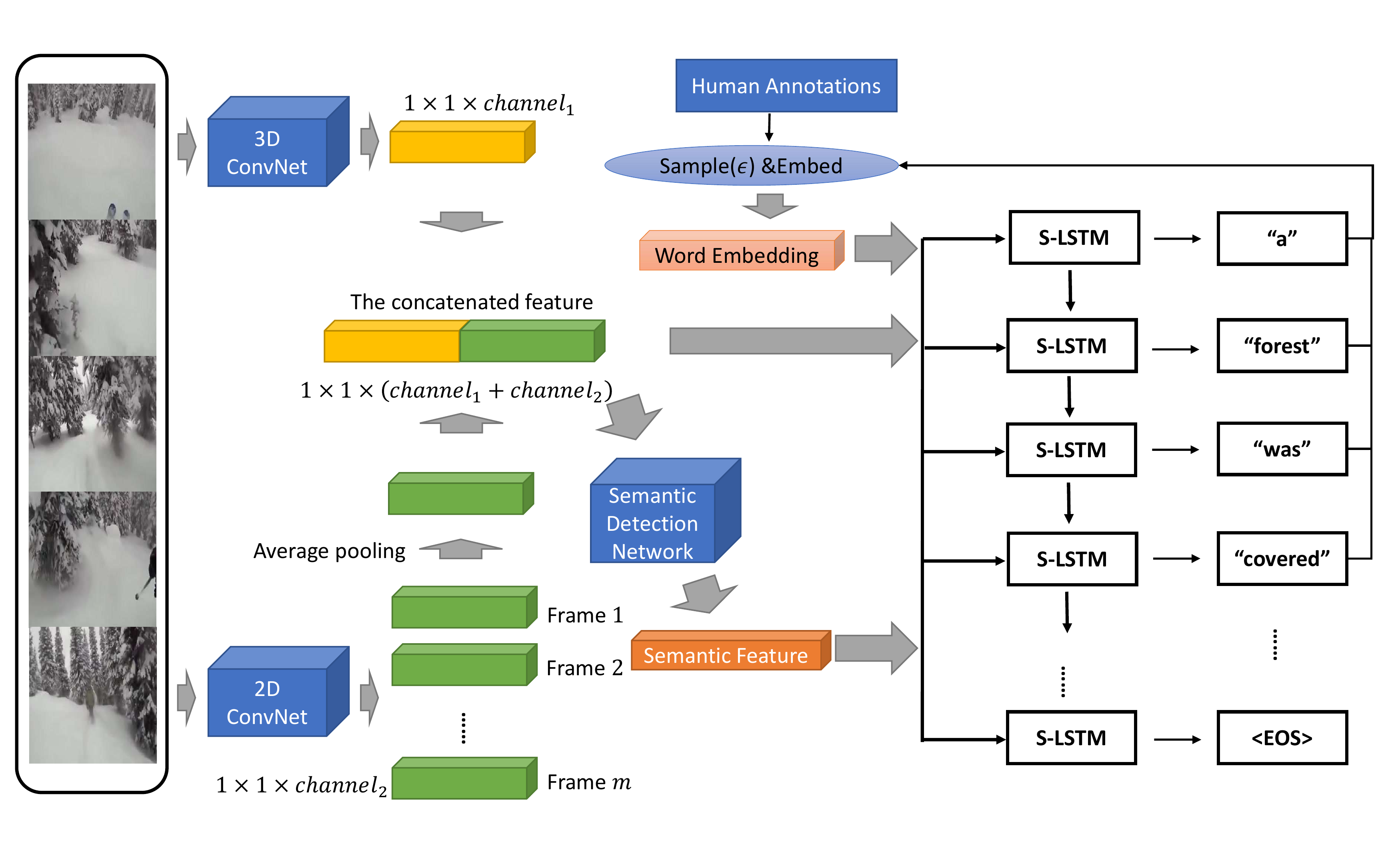}
\end{center}
\caption{Overall framework of our model. 
A 3D ConvNet, a 2D ConvNet and a Semantic Detection Net (SDN) constitute the encoder component of our model. 
S-LSTM stands for a semantics-assisted variant of LSTM which takes a semantic feature, 
a visual feature and a word embedding as inputs at each step. 
The word fed as the input to the decoder is sampled from human annotations or the model itself randomly, 
and then is embeded with the pretrained weights.}
\label{figure:f1}
\end{figure*}

The overall structure of our model is visualized in Figure \ref{figure:f1}. 
Our SDN and visual feature extractors in the encoder component share the same 2D ConvNet and 3D ConvNet in practice. 

\section{Experiments}
We evaluate our model on two popular video captioning datasets to show the performance of our approach. 
We compare our results to other existing methods.

\subsection{Datasets}
\subsubsection{YouTube2Text}
The YouTube2Text or MSVD \cite{guadarrama2013youtube2text, chen:acl11} dataset, published in 2013, 
contains 1970 short YouTube video clips. 
The average length of them is about 10 seconds. 
We get roughly 40 descriptions for each video. 
We follow the dataset split setting used in prior studies \cite{pan2016jointly, yu2016video, gan2017semantic}, 
in which the training dataset contains 1200 clips, the validation dataset contains 100 clips, 
and the rest of them belong to the test dataset. 
We tokenize the captions from the training and validation datasets and obtain approximately 14000 unique words. 
12592 of them are uitilzed for prediction, and the remaining words are replaced by $<unk>$. 
We add the token $<eos>$ to signal the end of a sentence.

\subsubsection{MSR-VTT}
MSR-Video to Text (MSR-VTT) \cite{xu2016msr-vtt, pan2016jointly} is a large-scale video benchmark, first presented in 2016. 
In its first version, MSR-VTT provided 10k short video segments with 200k descriptions in total. 
Each video segment was described by about 20 independent English sentences. 
In its second version, which was published in 2017, MSR-VTT provides additional 3k short clips as a testing set, 
and video clips in the first version can be used as training and validation sets. 
Because of lacking human annotations for the test set in the second version, we perform experiments on the first version. 
We tokenize and obtain 14071 unique words that appear in the training set and validation set of MSR-VTT 1.0 more than once. 
13794 of them are indexed with integer numbers starting at 0, and the rest are substituted by $<unk>$. 
$<eos>$, which signifies the end of a sentence, is added to the vocabulary of MSR-VTT.

\subsection{Overall Score}
Based on the widely used BLEU, METEOR, ROUGE-L and CIDEr metrics, 
we propose an overall score to evaluate the performance of a language model: 
\begin{equation}
\begin{aligned}
&\mathbf{S}_{overall}= \\ 
&\frac{\text{B-4}}{top1(\text{B-4})}+\frac{\text{C}}{top1(\text{C})}+
\frac{\text{M}}{top1(\text{M})}+\frac{\text{R}}{top1(\text{R})} \in [0,1], 
\end{aligned}
\label{eq:overall}
\end{equation}
where B-4 denotes BLEU-4, C denotes CIDEr, M denotes METEOR, R represents ROUGE-L 
and $top1(\cdot)$ denotes the best numeric value of the specific metric. 
We presume that BLEU-4, CIDEr, METEOR and ROUGE-L reflect one particular aspect of the performance of a model respectively. 
First, we normalize each metric value of a model, 
and then we take the mean value of them as an overall measurement for that model \eqref{eq:overall}. 
If the result of a model on each metric is closer to the best result of all models, the overall score will be close to 1. 
If and only if a model has the state-of-the-art performance on all metrics, the overall score is 1. 
If a model is much lower than the state-of-the-art result on each metric, the overall score of the model will be close to 0. 

\subsection{Training Details}
Our visual feature consists of two parts: a static visual feature and a dynamic visual feature. 
ResNeXt \cite{xie2017aggregated}, which is pretrained on the ImageNet ILSVRC2012 dataset, 
is utilized as the static visual feature extractor in the encoder of our model. 
The ECO \cite{zolfaghari2018eco}, which is pretrained on the Kinetics-400 dataset, 
is utilized as the dynamic visual feature extractor for the encoder in our model. 
More specifically, 32 frames are extracted from each video clip evenly. 
For each video, we feed 32 frames as input to ResNeXt, take the conv5/block3 output, 
and apply averge pooling to these outputs along the time axis. 
The newly obtained 2048-dim feature vector is taken as the 2D representation of that video.
What's more, we take the 1536-way feature of the global pool in ECO as the 3D representation of each video. 
We set the initial learning rate to $2 \times 10^{-4}$ for the YouTube2Text dataset and $4 \times 10^{-4}$ for the MSR-VTT dataset. 
In addition, we drop the learning rate by 0.316 every 20350 steps for the MSR-VTT dataset. 
Batch size is set to 64, and the Adam algorithm is applied to optimize the model for both datasets. 
The hyper-parameter $\beta_1$ is set to 0.9, $\beta_2$ is set to 0.999, and $\epsilon$ is set to $1\times10^{-8}$ for the Adam algorithm. 
Each model is trained for 50 epochs, 
in which the hyper parameter sample probability $\epsilon$ is set as $ep \times 0.008$ for the $ep$th epoch. 
We fine-tune the hyper-parameters of our model on the validation sets 
and select the best checkpoint for testing according to the overall score of the evaluation on the validation set. 

\subsection{Comparison with Existing Models}
Empirically, we evaluate our method on the YouTube2Text/MSVD \cite{guadarrama2013youtube2text} and MSR-VTT \cite{xu2016msr-vtt} datasets. 
We report the results of our model along with a number of existing models in Tables \ref{table:t1} and \ref{table:t2}.

\begin{table}[htb]
\centering
\begin{threeparttable}
\caption{Result comparison with existing models on the YouTube2Text dataset}\label{table:t1}
\begin{tabular} {llccccc}
\toprule
Model 											& B-4 		& C 		& M 		& R 		& Overall \eqref{eq:overall} \\
\midrule
  LSTM-E (V+C3D)  \cite{pan2016jointly} 	 	& 45.3 		& 			& 31.0 		& 			& \\
  h-RNN (V+C3D) \cite{yu2016video} 		 	& 49.9 		& 65.8 		& 32.6 		& 			& \\
  aLSTMs (I-3) \cite{gao2017video} 	 		& 50.8 		& 74.8 		& 33.3 		& 			& \\
  SCN (R-152+C3D) \cite{gan2017semantic} 		& 51.1 		& 77.7 		& 33.5		& 			& \\
  MTVC (I-4) \cite{pasunuru2017multi-task}		& 54.5 		& 92.4 		& 36.0 		& 72.8 		& 0.8961 \\
  ECO (R-152+E) \cite{zolfaghari2018eco} 		& 53.5 		& 85.8 		& 35.0 		& 			&\\
  SibNet (I-1) \cite{liu2018sibnet} 			& 54.2 		& 88.2 		& 34.8 		& 71.7 		& 0.8740 \\
  POS (IR+I3D) \cite{Wang_2019_ICCV}			& 53.9		& 91.0		& 34.9		& 72.1		& 0.8811 \\
  MARN (R-101+R3D) \cite{Pei_2019_CVPR}		& 48.6		& 92.2		& 35.1		& 71.9		& 0.8633 \\
  JSRL-VCT (IR+C3D) \cite{Hou_2019_ICCV}		& 52.8		& 87.8		& 36.1		& 71.8		& 0.8762 \\
  GRU-EVE (IR+C3D) \cite{Aafaq_2019_CVPR}		& 47.9		& 78.1		& 35.0		& 71.5		& 0.8264 \\
  STG-KD (R-101+I3D) \cite{Pan_2020_CVPR}		& 52.2		& 93.0		& 36.9		& 73.9		& 0.8975 \\
  SAAT (IR+C3D) \cite{Zheng_2020_CVPR}			& 46.5		& 81.0		& 33.5		& 69.4		& 0.8110 \\
  ORG-TRL (IR+C3D) \cite{Zhang_2020_CVPR}		& 54.3		& 95.2		& 36.4		& 73.9		& 0.9078 \\
 \midrule
  Our model 							& \textbf{62.4} 	& \textbf{109.7} & \textbf{39.0} 	& \textbf{77.0} & \textbf{1.0000}\\
\bottomrule
\end{tabular}
\begin{tablenotes}
\item[a] V, C3D, I-n, R-n, E, IR, I3D and R3D denote VGG19, C3D, n-version Inception, n-layer ResNet, ECO, Inception-ResNet-v2, I3D and 3D-ResNeXt features, respectively.
\end{tablenotes}
\end{threeparttable}
\end{table}

\begin{table}[htb]
\centering
\begin{threeparttable}
\caption{Result comparison with existing models on the MSR-VTT dataset}\label{table:t2}
\begin{tabular}{lccccc}
\toprule
 Model 												& B-4 		& C 			& M 			& R 			& Overall \\
 \midrule
 MTVC (I-4) \cite{pasunuru2017multi-task} 			& 40.8 		& 47.1 			& 28.8 			& 60.2 			& 0.9223 \\
 CIDEnt-RL (I-4) \cite{pasunuru2017reinforced} 	& 40.5 		& 51.7 			& 28.4 			& 61.4 			& 0.9435 \\
 SibNet (I-3) \cite{liu2018sibnet} 				& 40.9 		& 47.5 			& 27.5 			& 60.2 			& 0.9137 \\
 HACA (R-152+A\tnote{a})  \cite{wang2018watch} 				& 43.4 		& 49.7 			& 29.5 			& 61.8 			& 0.9608 \\
 TAMoE (I3D) \cite{wang2019learning} 				& 42.2 		& 48.9 			& 29.4 			& 62.0 			& 0.9505 \\
 POS (IR+I3D) \cite{Wang_2019_ICCV} 				& 41.3		& \textbf{53.4}	& 28.7			& 62.1 			& 0.9611 	\\
 MARN (R-101+R3D) \cite{Pei_2019_CVPR}				& 40.4		& 47.1			& 28.1			& 60.7			& 0.9162	\\
 JSRL-VCT (IR+C3D) \cite{Hou_2019_ICCV}			& 42.3		& 49.1			& \textbf{29.7}	& 62.8			& 0.9576	\\
 GRU-EVE (IR+C3D) \cite{Aafaq_2019_CVPR}			& 38.3		& 48.1			& 28.4			& 60.7			& 0.9119	\\
 STG-KD (R-101+I3D) \cite{Pan_2020_CVPR}			& 40.5		& 47.1			& 28.3			& 60.9			& 0.9192	\\
 SAAT (IR+C3D+Ca\tnote{a}) \cite{Zheng_2020_CVPR}			& 39.9		& 51.0			& 27.7			& 61.2			& 0.9303	\\
 ORG-TRL (IR+C3D) \cite{Zhang_2020_CVPR}			& 43.6		& 50.9			& 28.8			& 62.1			& 0.9628	\\
 \midrule 
 Our model 											& \textbf{45.8} & 53.2			& 29.3 			& \textbf{63.6} & \textbf{0.9957} \\
\bottomrule
\end{tabular}
\begin{tablenotes}
\item[a] A and Ca denote audio and category features, respectively.
\end{tablenotes}
\end{threeparttable}
\end{table}

\subsubsection{Comparison on the YouTube2Text Dataset}
Table \ref{table:t1} displays the performance of several models on YouTube2Text. 
We compare our model with existing methods, 
including LSTM-E\cite{pan2016jointly}, h-RNN\cite{yu2016video}, 
aLSTMs\cite{gao2017video}, SCN\cite{gan2017semantic}, MTVC\cite{pasunuru2017multi-task}, 
ECO\cite{zolfaghari2018eco}, SibNet\cite{liu2018sibnet}, POS\cite{Wang_2019_ICCV}, MARN\cite{Pei_2019_CVPR}, 
JSRL-VCT\cite{Hou_2019_ICCV}, GRU-EVE\cite{Aafaq_2019_CVPR}, STG-KD\cite{Pan_2020_CVPR}, 
SAAT\cite{Zheng_2020_CVPR} and ORG-TRL\cite{Zhang_2020_CVPR}. 
Our method outperforms all the other methods on all the metrics by a large margin. 
Note that many of them were published after our initial submission of the present work in the end of May in 2019. 
Specifically, compared with ORG-TRL \cite{Zhang_2020_CVPR}, the previous state-of-the-art model on this dataset, 
BLEU-4, CIDEr, METEOR and ROUGE-L are improved relatively by 14.9\%, 15.2\%, 7.1\% and 4.2\%, respectively.
Our model has the highest overall score as defined in \eqref{eq:overall}.

\subsubsection{Comparison on the MSR-VTT Dataset}
Table \ref{table:t2} displays the evaluation results of several video captioning models on the MSR-VTT. 
In this table, we compare our model with existing models, 
including MTVC\cite{pasunuru2017multi-task} , CIDEnt-RL\cite{pasunuru2017reinforced}, SibNet\cite{liu2018sibnet}, 
HACA\cite{wang2018watch}, TAMoE\cite{wang2019learning}, 
POS\cite{Wang_2019_ICCV}, MARN\cite{Pei_2019_CVPR}, JSRL-VCT\cite{Hou_2019_ICCV}, GRU-EVE\cite{Aafaq_2019_CVPR}	, 
STG-KD\cite{Pan_2020_CVPR}, SAAT\cite{Zheng_2020_CVPR}, ORG-TRL\cite{Zhang_2020_CVPR}. 
According to the overall score defined in \eqref{eq:overall}, ORG-TRL is the best among existing models. 
Our model achieves higher values on all metrics than this model. 
Two models POS and JSRL-VCT achieve slightly higher CIDEr value and METEOR values than our model, 
respectively, but their other metric values are clearly lower than our results.  

Our model achieves better results on both the YouTube2Text dataset and the MSR-VTT dataset. 
Note that our model is only trained on a single dataset without an attention mechanism, and it is tested without ensemble or beam search.

\section{Model Analysis}
In this section, we discuss the utility of the three improvements on our model. 

\subsection{Analysis on Semantic Features}
Semantic features are the output of a multi-label classification task. 
Mean average precision (mAP) is often used to evaluate the results of multi-label classification tasks \cite{tsoumakas2007multi-label}. 
Here, we apply it to evaluate the quality of semantic features. 

\subsubsection{Semantic Features Predicted with Different Sets of Input Features}
\begin{figure}[htb]
\centering
\includegraphics[width=0.45\textwidth]{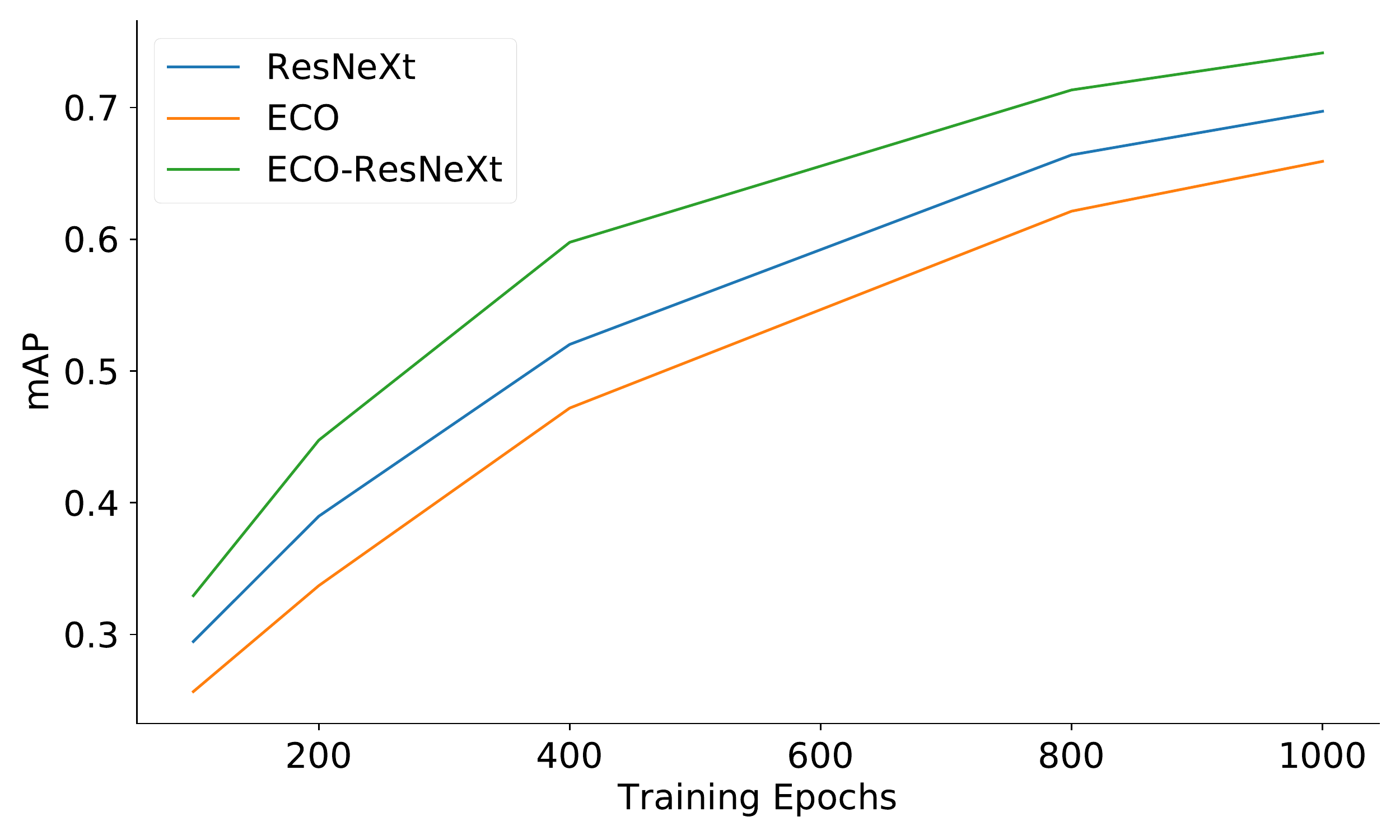}
\caption{The quality of semantic features predicted with different sets of input features evaluated 
by mAP on the YouTube2Text. 
"ResNeXt", "ECO" and "ECO-ResNeXt" denote that 
the semantic models are trained and the semantic features are predicted 
with visual features produced by ResNeXt, ECO, both ECO and ResNeXt, respectively. }
\label{figure:semanticsMSVD}
\end{figure}

\begin{figure}[htb]
\centering
\includegraphics[width=0.45\textwidth]{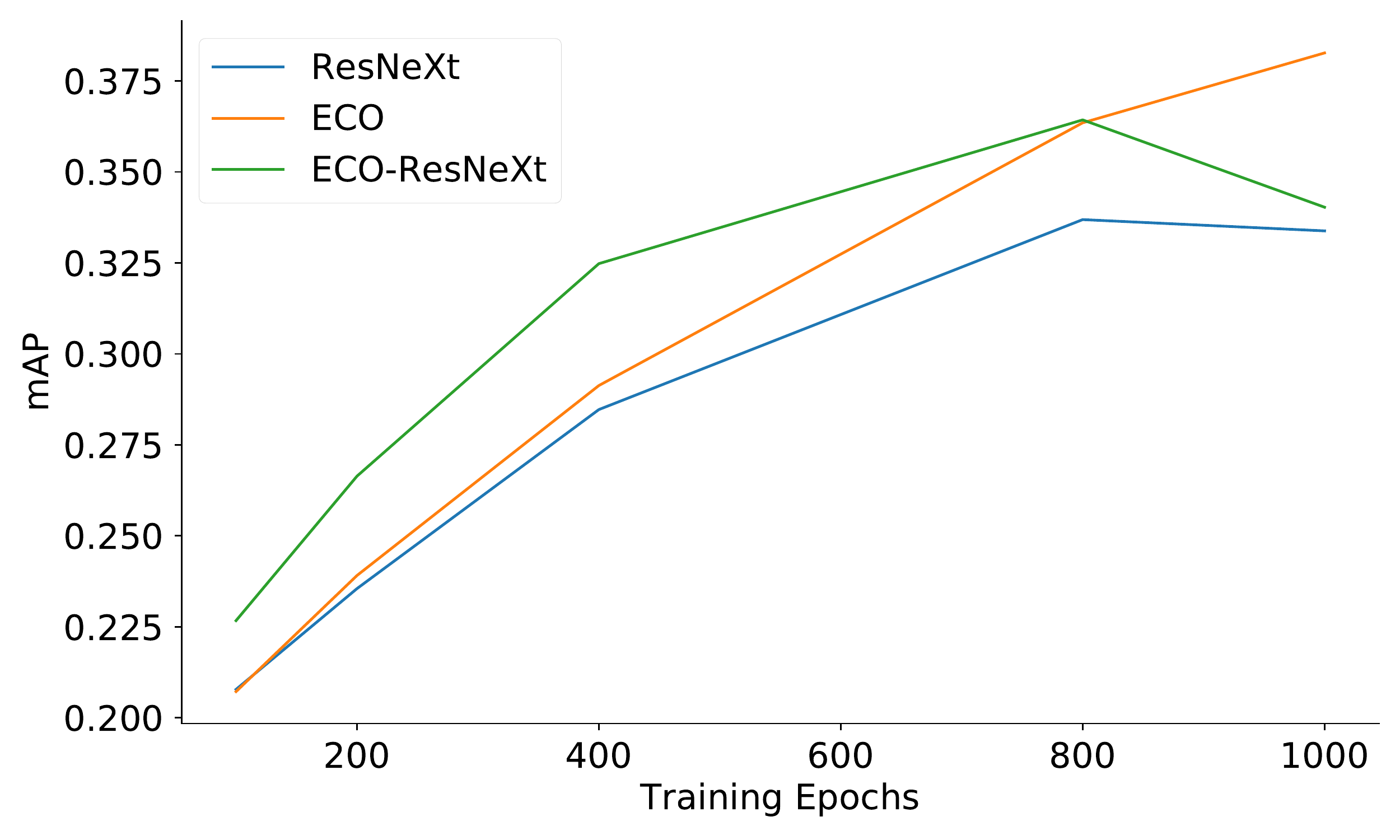}
\caption{The quality of semantic features predicted with different sets of input features evaluated 
by mAP on the MSR-VTT dataset.}\label{figure:semanticsMSRVTT}
\end{figure}

Figures \ref{figure:semanticsMSVD} and \ref{figure:semanticsMSRVTT} demonstrate the quality of semantic features, 
using different sets of feature maps as inputs, with respect to the training epochs. 
Figure \ref{figure:semanticsMSVD} shows that, on the YouTube2Text dataset, 
the mAP values are proportional to training epochs. 
With the same number of training epochs, the qualities of semantic features are in the order: ECO-ResNeXt $>$ ResNeXt $>$ ECO, 
where ECO-ResNeXt, ResNeXt and ECO denote the models trained with visual features from ECO-ResNeXt, ResNeXt or ECO, respectively. 
Figure \ref{figure:semanticsMSRVTT} demonstrates that, on the MSR-VTT dataset, both mAP values of semantic information 
decline after the models are trained for more than 800 epochs with ResNeXt feature maps or ECO-ResNeXt feature maps as inputs. 
With ECO feature maps as inputs, the performance of the semantic detection model is still proportional to the training epochs. 

\subsubsection{Models Trained with Different Semantic Features}
Tables \ref{table:t3} and \ref{table:t4} list the performance of our model trained by scheduled multinomial sampling with different semantic features on the YouTube2Text and MSR-VTT datasets, respectively. 
The results clearly show that a better multi-label classification enables a better video captioning model. 
Semantic features with higher mAP provide more appropriate potential attributes of a video for the model. 
Thus, the model is able to generate better video annotations by comprehensively considering semantic features, spatio-temporal features and contextual information. 
\begin{table}[htb]
\centering
\begin{threeparttable}
\caption{Results of scheduled sampling methods (multinomial sampling) on the YouTube2Text dataset with different sets of semantic features}\label{table:t3}
\begin{tabular}{cccccc}
\toprule
Semantic Features (mAP\tnote{a}) 		& B-4 				& C 				& M 			& R 				& Overall \\
\midrule
0.3295 							& 53.9 				& 90.5 				& 35.8 			& 73.4 				& 0.8896 \\
0.5977 							& 60.5 				& 102.7 			& 38.0 			& 75.9 				& 0.9663 \\
0.7414 							& \textbf{62.4} 	& \textbf{109.7} 	& \textbf{39.0} & \textbf{77.0} 	& \textbf{1.0000} \\
\bottomrule
\end{tabular}
\begin{tablenotes}
\item[a] A larger mAP implies a better representation of semantic meanings
\end{tablenotes}
\end{threeparttable}
\end{table}

\begin{table}[htb]
\centering
\caption{Results of scheduled sampling methods (multinomial sampling) on MSR-VTT data with different sets of semantic features}\label{table:t4}
\begin{tabular}{cccccc}
\toprule
Semantic Feature (mAP) 		& B-4 				& C 				& M 				& R 				& Overall \\
\midrule
0.2072 						& 40.5 				& 46.8 				& 27.2 				& 62.7 				& 0.9292 \\
0.2913 						& 44.0 				& 50.7 				& \textbf{28.9} 	& 62.6 				& 0.9878 \\
0.3827 						& \textbf{44.9} 	& \textbf{51.8} 	& 28.8 				& \textbf{63.12} 	& \textbf{0.9996} \\
\bottomrule
\end{tabular}
\end{table}
\subsection{Analysis on the Scheduled Sampling}
Tables \ref{table:t5} and \ref{table:t6} show the comparison among the Teacher Forcing algorithm, 
scheduled sampling with the $\arg\max$ strategy and scheduled sampling with the multinomial strategy on YouTube2Text and MSR-VTT datasets, respectively. 
Teacher Forcing utilizes human annotations to guide the generation of words during training and samples from the word distribution of the output of the model to direct the generation during inference. 
The $\arg\max$ strategy switches gradually from the Teacher Forcing way to sample words with the largest possibility from the model itself during training. 
The Multinomial strategy is similar to the $\arg\max$ strategy but samples words randomly from the distribution of the model at each step. 
As we can infer from Tables \ref{table:t3} and \ref{table:t4}, 
the scheduled sampling with the multinomial strategy yields a better performance than the other two methods on the YouTube2Text dataset and the one with the $\arg\max$ strategy yields the best performance on the MSR-VTT dataset. 
Our method explores a larger range of RNN state space and thus is likely to find a better solution during training.

\begin{table}[htb]
\centering
\caption{Results of different training strategies on YouTube2Text data with the best semantic features}\label{table:t5}
\begin{tabular}{cccccc}
\toprule
Training Method & B-4 				& C 				& M 				& R 				& Overall \\
\midrule
Teacher Forcing & 61.93 			& 108.56 			& 38.96 			& 76.75 			& 0.9942 \\
$\arg \max$ 	& 62.16 			& 109.31 			& 38.98 			& 76.81 			& 0.9972 \\
Multinomial 	& \textbf{62.35} 	& \textbf{109.71} 	& \textbf{39.04} 	& \textbf{77.04} 	& \textbf{1.0000} \\
\bottomrule
\end{tabular}
\end{table}

\begin{table}[htb]
\centering
\caption{Results of different training strategies on MSR-VTT data with the best semantic features}\label{table:t6}
\begin{tabular}{cccccc}
\toprule
Training Method & B-4 				& C 				& M 				& R 				& Overall \\
\midrule
Teacher Forcing & 45.05 			& 50.25 			& 29.12 			& 62.72 			& 0.9771 \\
$\arg \max$    	& \textbf{45.83} 	& \textbf{53.16} 	& \textbf{29.28} 	& \textbf{63.64} 	& \textbf{1.0000} \\
Multinomial  	& 44.94 			& 51.77 			& 28.82 			& 63.12 			& 0.9826 \\
\bottomrule
\end{tabular}
\end{table}

\subsection{Analysis on the Length Normalization of the Loss Function}
\begin{figure*}[htb]
\centering
\begin{tabular} {ll}
\toprule \hline
\begin{minipage}{0.32\textwidth}
\includegraphics[height=0.09\textheight,width=1\textwidth]{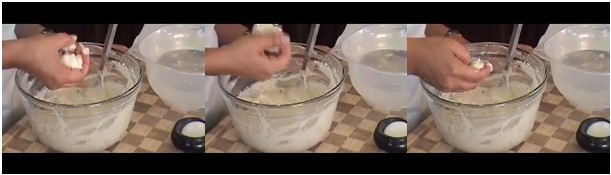}\end{minipage} &  
\makecell[lc]{\\
$\beta=0   $:   a woman is mixing a bowl\\
$\beta=0.7 $:   a woman is mixing a bowl\\
$\beta=1   $:   a person is mixing a bowl of a bowl\\
GT:  somebody is mixing flour\\\\ 
}\\ \hline
\begin{minipage}{0.32\textwidth}
\includegraphics[height=0.09\textheight, width=1\textwidth]{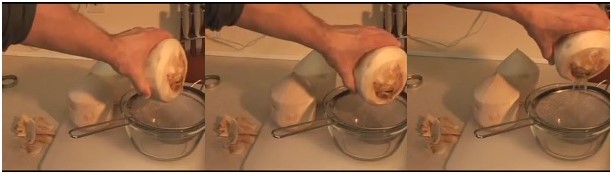}\end{minipage} & 
\makecell[lc]{\\
$\beta=0   $:   a man is pouring a egg\\
$\beta=0.7$:   a man is pouring eggs into a bowl\\
$\beta=1   $:    a man is adding a bowl of a bowl\\
GT: a man is pouring coconut juice into a bowl\\\\
}\\ \hline
\begin{minipage}{0.32\textwidth}
\includegraphics[height=0.09\textheight, width=1\textwidth]{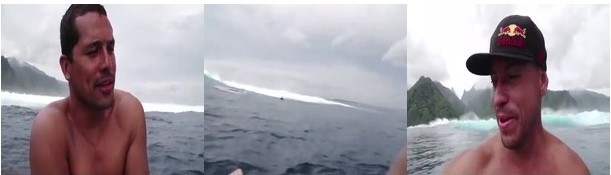}\end{minipage} & 
\makecell[lc]{\\
$\beta=0   $:    a man is talking about a boat\\
$\beta=0.7$:   a main is talking about the water\\
$\beta=1   $:     a man is talking about the the the the the the the the \\
GT: some men having fun and talking about the sea\\\\
}\\ \hline
\begin{minipage}{0.32\textwidth}
\includegraphics[height=0.09\textheight, width=1\textwidth]{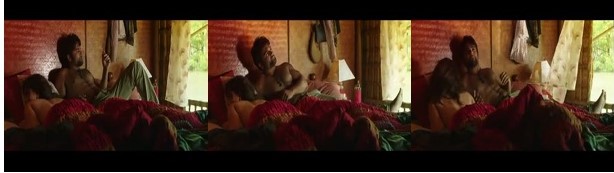}\end{minipage} & 
\makecell[lc]{\\
$\beta=0   $:    a woman is sitting on a couch\\
$\beta=0.7$:   a man and a woman are sitting in a bed\\
$\beta=1   $:     a man is sitting on a bed and a woman is sitting on the bed \\
GT: a man and woman are lying in bed together\\\\
}\\ \hline
\bottomrule
\end{tabular}
\caption{Examples of machine-generated captions and human annotations(GT).}
\label{figure:figtab}
\end{figure*}
As demonstrated in Table \ref{table:t7}, the average length of human annotations is larger than those generated by models with $\beta=\{0, 0.7, 1\}$ \eqref{eq:loss1}, respectively. 
But Figure \ref{figure:figtab} displays the tendency of redundancy in captions generated by the $\beta=1$ model, which deteriorates the overall quality of model-generated sentences. 
The average caption length of the model with $\beta=0.7$ is greater than that of the model with $\beta=0$, whereas it is smaller than that from the model with $\beta=1$. 
The model with $\beta=0.7$ generates relatively long annotations for videos without suffering from redundancy or duplication of words, and we therefore consider it the optimal choice.
\begin{table}[htb]
\centering
\begin{threeparttable}
\caption{Average length of the captions in the test set}\label{table:t7}
\begin{tabular}{ccccc}
\toprule
Model & $\beta=0$ 	& $\beta=0.7$ 	& $\beta=1$ 	& Ground Truth\tnote{b} \\
\midrule
mLen1\tnote{a} & 5.12 		& 5.18 			& 5.80   		& 7.01 \\
mLen2\tnote{a} & 6.27 		& 6.69 			& 6.99   		& 9.32 \\
\bottomrule
\end{tabular}
\begin{tablenotes}
\item[a] mLen1 stands for the mean length of YouTube2Text, and mLen2 stands for the mean length of MSR-VTT. 
\item[b] Ground Truth denotes the human annotations for the test set.
\end{tablenotes}
\end{threeparttable}
\end{table}

\section{Conclusion}
We suggest three improvements for solving the video captioning task. 
First, mAP is applied to evaluate the quality of semantic information, 
and a SDN with adequate computation complexity and input features is used to extract high-quality semantic features from videos, 
which contributes to the success of our semantics-assisted model. 
Second, we employ a scheduled sampling training strategy. 
Third, a sentence-length-modulated loss function is proposed to keep the model in a balance between language redundancy and conciseness. 
Our method achieves results that are superior to the state-of-the-art on the YouTube2Text dataset. 
The performance of our model is comparable to the state-of-the-art on the MSR-VTT dataset. 
In the future, we may obtain further improvements in video captioning by integrating spatio-temporal attention mechanisms with visual-semantics features.

\appendices
\section*{Funding}
This work was supported in part by the National Key Research and Development Program of China under Grant 2017YFA0700904, in part by the National Natural Science Foundation of China under Grant Grant 61621136008, in part by the German Research Council (DFG) under Grant TRR-169, and in part by Samsung under contract NO. 20183000089.

\section*{Acknowledgment}
The authors thank Han Liu, Hallbjorn Thor Gudmunsson and Jing Wen for valuable and insightful discussions.

\ifCLASSOPTIONcaptionsoff
  \newpage
\fi

\bibliographystyle{IEEEtran}
\bibliography{./jrnl}

\end{document}